\documentclass[runningheads,a4paper]{llncs}
\usepackage{amssymb}
\usepackage{graphicx}
\usepackage[utf8]{inputenc}
\usepackage{url}
\usepackage{courier}
\usepackage{listings}
\newcommand{\keywords}[1]{\par\addvspace\baselineskip
\noindent\keywordname\enspace\ignorespaces#1}
\begin{document}

\mainmatter

\title{Contextual Data Collection for Smart Cities}

\author{Henrique Santos\inst{1,2}\and Vasco Furtado\inst{2,3}\and Paulo Pinheiro\inst{1}\and\\ Deborah L. McGuinness\inst{1}}
\authorrunning{Henrique Santos et al.}

\institute{Rensselaer Polytechnic Institute, Troy, NY, U.S.A.
\and
Universidade de Fortaleza, Fortaleza, CE, Brazil
\and
Fundação de Ciência, Tecnologia e Inovação de Fortaleza, Fortaleza, CE, Brazil}

\toctitle{Contextual Data Collection for Smart Cities}
\tocauthor{H. Santos, V. Furtado, P. Pinheiro and D. L. McGuinness}
\maketitle

\begin{abstract}
As part of Smart Cities initiatives, national, regional and local governments all over the globe are under the mandate of being more open regarding how they share their data. Under this mandate, many of these governments are publishing data under the umbrella of open government data, which includes measurement data from city-wide sensor networks. Furthermore, many of these data are published in so-called data portals as documents that may be spreadsheets, comma-separated value (CSV) data files, or plain documents in PDF or Word documents. The sharing of these documents may be a convenient way for the data provider to convey and publish data but it is not the ideal way for data consumers to reuse the data. For example, the problems of reusing the data may range from difficulty  opening a document that is provided in any format that is not plain text, to the actual problem of understanding the meaning of each piece of knowledge inside of the document. Our proposal tackles those challenges by identifying metadata that has been regarded to be relevant for measurement data and providing a schema for this metadata. We further leverage the Human-Aware Sensor Network Ontology (HASNetO) to build an architecture for data collected in urban environments. We discuss the use of HASNetO and the supporting infrastructure to manage both data and metadata in support of the City of Fortaleza, a large metropolitan area in Brazil.
\end{abstract}

\section{Introduction}

Smart Cities should be sustainable, safe, inclusive, walkable, creative and innovative. In one way or another, the availability of each one of these features characterize a desirable place to live. Obtaining any of them, however, requires two key components: access to and understanding of city data. Consequently, a city’s ability to produce and share relevant data that can be understood is critical and can be viewed as key indicator of a Smart City.  Further a city’s ability to derive knowledge from this data and further, to use it to power innovation is an even better indicator of a smart city.

Two complementary trends are particularly relevant in this context: the Internet of Things (IoT) and the Open Government Data approach (OGD). The IoT is already a reality in many large metropolitan areas around the world. A myriad of sensors operating at different levels of autonomy are deployed throughout cities collecting data from every aspect of these cities' rich urban environments. Governments are increasingly sharing their data, often with the goal of promoting innovation via societal participation with the use of the data. In the context of data sharing, different categories of stakeholders may be identified: designers and software developers may use data to produce public services through the use of web and mobile applications; scientists may produce elaborate analyses and studies about the cities; public officers may use the data to improve city administration through the use of effective data-based decision-making techniques; journalists may use open data to produce more reliable, factually-based and attractive news. Briefly, IoT and OGD are enabling  knowledge production that is fundamental for enhancing city “smartness”.  While foundational elements exist with IoT and OGD, many challenges remain to truly realize the potential of widespread knowledge generation from the raw data sources increasingly available in many cities.

One key challenge that we address in this paper is the need for cities to provide metadata that enables data consumers to understand publicly shared data, including the context within which data was collected. The lack of proper metadata significantly impacts the extraction of knowledge from data for any of stakeholders identified above. The reliability of data collected by extensive sensor networks, for example, relies heavily on knowledge about sensor deployment, sensor calibration, measurement units, measurement accuracy, and many other types of knowledge that is often lost at measurement time.

Open government data (which includes monitored data from city sensors) is typically published in data portals that provide access to the contents of datasets. It is true that datasets are a convenient way to convey and publish data but, while that holds true on a technological perspective, they are not always the best suited for final users to use as the boundaries of a dataset may be more driven by technical considerations of sensor collection rather than boundaries more natural for users.  Further the context for the boundaries is often not captured or communicated. The datasets are merely enclosures for data, which may be serialized inside them in a number of formats using different domain vocabularies. The datasets hopefully come with metadata that attempt to explain the formats used to serialize the data and also to give meaning to the domain vocabularies used.  Accurately interpreting the contents of the data in one data portal even with a metadata file is often a challenge and greater challenges exist when users attempt to integrate data from multiple portals.

Our work addresses the lack of a vocabulary to describe the knowledge about sensor network measurements in the context of Smart Cities. We introduce the Human-Aware Sensor Network Ontology for Smart Cities (HASNetO-SC) used to describe knowledge associated with the collection of city empirical data, focusing in particular on the collection of city empirical data coming from city-wide sensor networks. The rest of the paper is organized as it follows: In Section 2, we discuss work related to the handling of urban stream data and to the development of ontologies describing empirical data collection. In Section 3, we present and discuss our research choices regarding HASNetO-SC and the process of collecting urban data. In Section 4, we describe the use of a HASNetO-SC-based tool when applied to a network of sensors deployed throughout a large metropolitan area. In Section 5, we discuss our findings and future work.

\section{Related work}

\subsection{Open Data Portals}

Open government data initiatives around the world are often very large projects in terms of complexity and data volume \cite{hendler_us_2012} \cite{shadbolt_linked_2012}. Goals behind these initiatives range from government transparency to fostering societal engagement in government decisions. Typically open government data efforts for disseminating data are centered around the development and publication of online data portals. For example, many of these portals are built on top of CKAN\footnote{http://ckan.org}, which operates as a comprehensive dataset repository with data coming from a plethora of distinct government agencies. Many times, CKAN-hosted datasets convey data gathered empirically by city sensors. The data often are consolidated into datasets because of technology restrictions. This approach, although convenient for data producers,  is often a hindrance for data consumers.

Much work exists on the integration and publication of urban data collections. QuerioCity \cite{lopez_queriocity:_2012} describes a Linked Data platform to publish, search and link city data from static datasets or stream data coming from deployed sensors. This platform is able to create a semantic catalog that describes the content of the datasets that reuses standard vocabularies to improve interoperability and discoverability. AECIS \cite{gao_semantic_2014} describes an automated discovery and integration system for urban data streams. Both works, although capable of dealing with urban data streams, are not concerned with data quality, in terms of comprehensive metadata and the sensor network and context where the data was collected.

\subsection{Semantic sensor network and Observational data}

The concept of Observational data is treated in the literature \cite{quine_stimulus_1995} \cite{stasch_stimulus-centric_2009} \cite{probst_ontological_2006} as data that is obtained while sensing some property of an entity from the real world. The result of an observation is a value for that property \cite{usbeck_combining_2014}. City-wide sensors are also performing observations, making content annotation crucial. Annotations enable interoperability and discoverability, making data easier to be understood and thus (re)used. To leverage the potential benefits of data reuse, several approaches exist to model the infrastructure that generates the data and to describe data content and context.

Semantic Sensor Networks make use of the description of instruments and detectors (many times referred to as sensors in the literature) to leverage and maintain complex networks of sensors, while providing integration of the collected data. In \cite{compton_survey_2009}, twelve different sensor network ontologies are studied and compared. This work preceded the W3C’s Semantic Sensor Network Ontology (SSN) \cite{compton_ssn_2012}. SSN is an ontology that aims to describe sensors, observations and its related concepts, like sensor capabilities, measurement processes and deployments. SSN is able to annotate data in a manner that makes it possible to to tell if it is coming from a certain sensor, using some measurement process to measure a certain property of an entity of interest. BOnSAI \cite{stavropoulos_bonsai:_2012} and the SESAME Meter Data Ontology (SMDO) \cite{fensel_sesame-s:_2012} are other sensor network ontologies that are focused on smart buildings. Although they are all able to describe the sensor network behind the collected data, SSN does not rely on standard provenance approaches, like the W3C’s PROV-O. Besides that, SSN is an ontology, and is not concerned about how the data is conveyed from its collection in terms of format and datasets. BOnSAI and SMDO are not scientific-centric ontologies, not containing key concepts like scientific activities.

O\&M \cite{cox_observations_2011} is an XML implementation from the Open Geospatial Consortium(OGC) that defines a schema for modeling observations and their results. In \cite{kuhn_functional_2009}, an observation and measurement ontology is proposed that makes use of O\&M's definitions. OBOE (The Extensible Observation Ontology) \cite{madin_ontology_2007} is an ontology for ecological observational data. It provides a data model that captures measurement semantics and that can be used to streamline data integration. To achieve this goal, the OBOE ontology contains concepts and relationships for describing observational datasets.

The Human-Aware Sensor Network Ontology (HASNetO) \cite{pinheiro_hasneto_2015} is an ontology for describing the scientific activities involved with the data collection of observational data, i.e., data that is collected while sensing the environment. HASNetO reuses three ontologies to achieve this task: W3C’s PROV-O, the Virtual Solar-Terrestrial Observatory - Instrument module (VSTOI) \cite{fox_ontology-supported_2009} and OBOE. By using PROV-O, HASNetO is capable of asserting the provenance of the measured data using PROV terms: activities, entities and agents. HASNetO links VSTOI and OBOE concepts to PROV concepts: (i) vstoi:Deployment and oboe:Observation are subclasses of prov:Activity; (ii) vstoi:Dataset is a prov:Entity and (iii) vstoi:Instrument becomes a prov:Agent. By doing this, HASNetO enables provenance tracking of data collection activities using W3C’s recommended standard.

Collected data can be encoded in many distinct formats including CSV, XML, and NetCDF \cite{rew_netcdf:_1990}. In many cases, CSV is a format of choice because of its ease of use by either computers or people. People often manually enter collected data in a spreadsheet application (like MS Excel or LibreOffice Calc).  Spreadsheets are also capable of exporting content in CSV format. Basically, the CSV format can be seen as a minimalist enabling approach for data interoperability.

Regardless of the format, no single encoding provides effective mechanisms for annotating observational data in a way that supports observation as a contextualized measurement collection. For instance, CSV lacks features for expressing the semantics associated with the data contained in it, so it is challenging to know, in an automated and interoperable way, the meaning of the data enclosed inside a CSV file. For example, it can be difficult to determine if two entries are observationally equivalent (measured under the same conditions, using the same units, in the same area, etc.). Nonetheless, different agents may generate data in different formats and standards, making CSV even more difficult to process automatically.

Although there are existing approaches for accessing CSV metadata and also for providing a metadata vocabulary for CSV data, they are typically more concerned with content restrictions, rather than the context in which the CSV data was collected. W3C's drafts from the CSV on the Web Working Group\footnote{http://www.w3.org/2013/csvw/wiki/Main\_Page} elaborate on techniques for enabling the access of CSV metadata by describing the content metadata in a separate JSON file that makes use of RDF vocabulary. To bridge this gap, we propose Contextualized CSV (CCSV)\footnote{http://tw.rpi.edu/web/project/JeffersonProjectAtLakeGeorge/download/ccsv} as a format that deals with both content and context restrictions of the observational data enclosed in it. The CCSV dataset is basically a regular CSV file with a Turtle preamble on top of it. The Turtle is used to assert that a certain dataset comes from a particular instrument during an specific deployment, as context metadata. It is also used to assert content, by stating the property being measured and which column contains the measured value.

\section{A streamlined process to collect and publish data for Smart Cities}

Our approach relies on the HASNetO ontology. HASNetO was primarily conceived to be used in scientific environments by scientists who conduct observation and data collection activities on entities of interest, such as a physical, chemical, biological, social or cultural feature. While comprehensive enough to provide what scientists need to keep track of their activities, HASNetO faces challenges when dealing with data collected in large complex settings with broad user bases, such as urban environments. One of those challenges is that the collected data must be used not only by the city administration or people involved with their associated measurements (people who have some knowledge about the data at some level), but, most importantly, by citizens who have, in general, little to no knowledge about the collection environment and the collected data. For instance, the accuracy or resolution of an instrument used to collect data may not be as important for a regular citizen as it is for a scientist. To address these challenges, we propose a modified version of HASNetO called the Human-Aware Data Collection ontology for Smart Cities (HASNetO-SC). Our main goal with HASNetO-SC is to provide smart city stakeholders with a streamlined way of collecting, preserving and disseminating urban data with an appropriate level of contextual metadata for data understanding. The following subsections discuss the aspects of this process.

\subsection{HASNetO-SC}

In 2007, a report on the ranking of the smartest medium-sized European cities was published \cite{giffinger_smart_2007}. This report aimed to, among other goals, define the general concept of a Smart City by specifying a set of factors and indicators that a city should be pursuing in order to increase its level of “smartness”. The work presented six key areas (characteristics) that were gathered from a literature search on previous definitions of the term Smart City: smart economy, smart people, smart governance, smart mobility, smart environment and smart living. Inside each characteristic, a number of factors were also compiled to support it and, in  turn, each factor consisted of a set of indicators that can identify well a city is doing with respect to that factor.  We developed an extension to HASNetO, called HASNetO-SC, to describe concepts that exist in urban environments. HASNetO-SC has been motivated by the smart cities characteristics identified in the report (not to be mistaken with characteristics of physical entities), while focusing on the four that had higher potential to produce data that can be collected empirically and managed using a streamlined process:

\begin{itemize}
\item Smart people
\item Smart mobility
\item Smart environment
\item Smart living
\end{itemize}

The extension basically refines VSTOI concepts that are integrated into HASNetO to better suit urban data collections. The following subsections further detail our modeling and discuss how we are using HASNetO concepts to cope with smart cities.

\subsubsection{People}

People are a central component of cities.  Cities are made for the citizens, as the government is intended to foster its society as a whole. Citizens, use city facilities and participate in the government discussions. In this scenario, people are capable of providing valuable information to the city. In previous work, \cite{goodchild_citizens_2007} has studied the view of a citizen as a sensor. Our approach is compatible with this perspective and enables people to either deploy instruments or conduct data collection activities. By the means of prov:Person, a person is a prov:Agent , as asserted in the PROV ontology, and the person can have an associated list of prov:Activity. In HASNetO, prov:Activity is extended to both vstoi:Deployment and hasneto:DataCollection.

\subsubsection{Mobility}

\begin{figure}
\centering
\includegraphics[height=7cm]{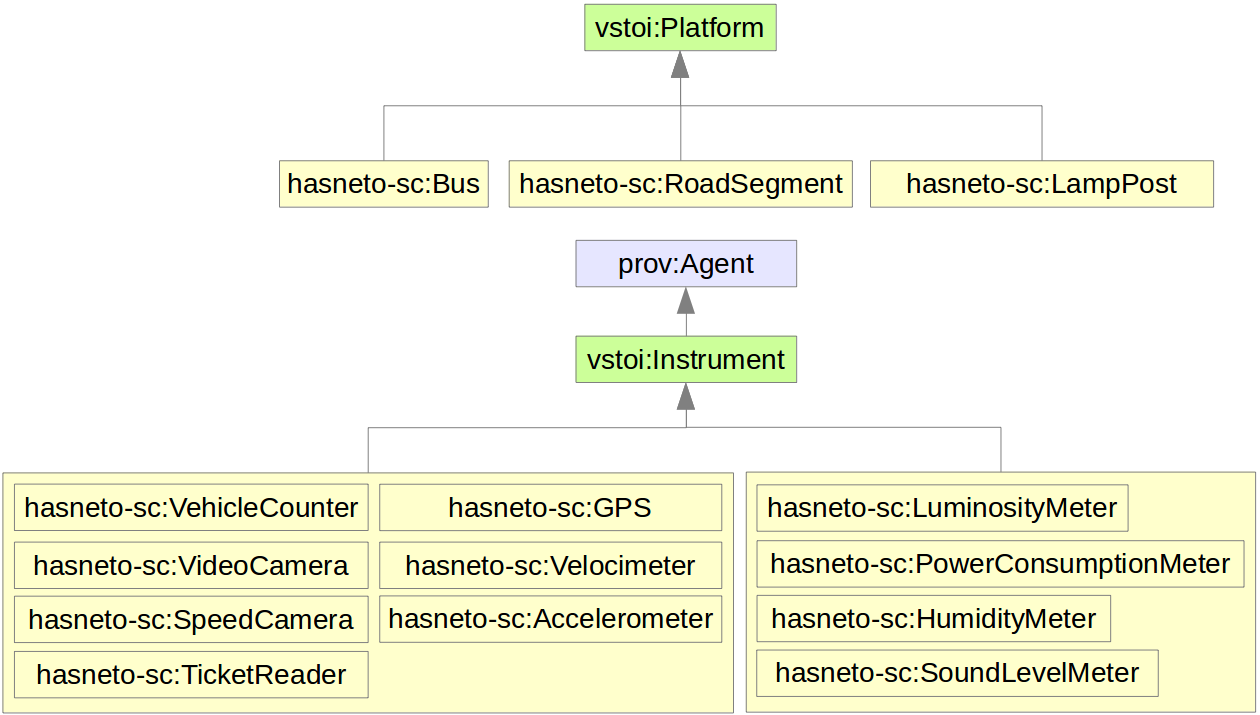}
\caption{HASNetO-SC classes.}
\label{hasneto-sc}
\end{figure}

Mobility is one of the most discussed and explored aspects of smart cities, due to its need of near-real time raw data (i.e., data from sensors) and derived data (i.e., results of data analysis). Mobility is central to many key city goals including accessibility, safe and sustainable transportation etc. To achieve those goals, many city governments are deploying sensors to monitor a number of urban mobility indicators. HASNetO enables provenance capture by describing how sensors are deployed, in particular how agents perform instrument deployments at platforms. In VSTOI, a Platform is a surface where Instruments and Detectors may be attached to measure, in the context above, characteristics of urban mobility. As a mobility example, we can identify some platforms and instruments that are known to be related to urban mobility. We have extended both vstoi:Platform and vstoi:Instrument to describe these more specialized mobility-related classes, as seen in Fig.~\ref{hasneto-sc}. The figure shows that vstoi:Platform includes three classes: subclasses hasneto-sc:Bus, hasneto-sc:RoadSegment, and hasneto-sc:LampPost. In a city, we defined buses, roads and lamp posts as being capable of hosting mobility-related instruments. Furthermore, vstoi:Instrument has been extended to a series of generic instruments that we propose should be used as a basis from which to derive specialized instruments, as each city will have distinct manufacturers, suppliers and vendors of instruments and platforms. The generic modeling approach described above makes it possible to further connect data and metadata, making them meaningful to data consumers (e.g., citizens) who are not necessarily involved with the specification of specialized instruments, even though they may know consumer instruments including cameras and GPSs.

\subsubsection{Environment}

Another well-discussed smart city aspect is environment. Examples of smart environment entities are “air”, “water”, and their associated monitoring and governance efforts such as “sustainable resource management”, “pollution control”,  and “environmental protection”. Forward-thinking cities are often concerned with citizen well being and thus are critically interested in environmental aspects. Signals of efficient use of water, electricity, and green space are some examples of indicators that a city that is doing well in the environment characteristic. To evaluate those indicators, many instruments are already in use in big cities that enable the collection of  data about air quality, noise levels, water conditions and so forth. Fig.~\ref{hasneto-sc} shows more specialized platforms and instruments. Again, those concepts are generic concepts that should be specialized to fully describe the actual city sensor network.

\subsubsection{Living}

Smart living includes the notions of public health, public safety, education and cultural facilities. Smart buildings are becoming a reality with more sensors being deployed to control access and gather the people flow information. Public services addressing public health and safety considerations often are placing sensors including cameras to help monitor potentially dangerous situations. Fig.~\ref{hasneto-sc} also depicts some of the most common platforms and instruments relevant to smart city living.

\subsection{Contextualized Comma Separated Values}

As mentioned before, we proposed CCSV as an extension to CSV to address both content and context restrictions when dealing with observational data. In our approach, both content and context metadata are needed in order to provide the connection between data and metadata. From the Smart City perspective, we needed to align the Turtle preamble of a CCSV dataset with the ontology concepts introduced by HASNetO-SC. Further in this subsection, the displayed code is an example of a CCSV preamble of a city dataset.

The preamble contains the following descriptions:
\begin{itemize}
\item Knowledge base: In order to provide the possibility of multi-contextual data collections and also to make the solution more scalable, we make it possible for the dataset to state which knowledge base should be used for validation and persistence.
\item Deployment: Deployment information makes it possible to link the data the CCSV dataset conveys to metadata information: (i) instrument and detector used; (ii) platform and (iii) all attached information to those, including accuracy, precision, platform location etc.
\item Data collection: Data collection is a prov:Activity as asserted in HASNetO. The use of Data collection information in the CCSV preamble enables the architecture to have knowledge about this dataset: it is able to know if distinct datasets were produced under the same context. In other words, the architecture is able to provide the user enough context information for him to decide if the data within different datasets can be comparable, joined, analyzed together, based on his needs.
\item Dataset: As previously discussed, datasets are not scientific boundaries for data, but data collection activities are. In this description, we link datasets to their corresponding data collections.
\item Measurement(s): Here all measured characteristics are described. For each measurement type we link its description to a unit and a measured characteristic. We’ll discuss the need for a domain ontology in the next section.
\end{itemize}

\medskip

\noindent
{\it Example of a CCSV Preamble}
\begin{lstlisting}[frame=single]
<city-kb>
  a ccsv:KnowledgeBase;
  ccsv:hasConnectionURL [some_url]^^xsd:anyURI .

<deployment>
  a vstoi:Deployment;
  prov:startedAtTime [some_timestamp]^^xsd:dateTime;
  hasneto:hasDataCollection <data-collection-001> .

<data-collection>
  a hasneto:DataCollection;
  a time:Interval;
  prov:startedAtTime [some_timestamp]^^xsd:dateTime .

<dataset>
  a vstoi:Dataset;
  prov:wasGeneratedBy <data-collection>;
  hasneto:hasMeasurementType <mt0> .

<mt0>
  a oboe:Measurement;
  a time:Instant;
  time:inDateTime <ts0>;
  ccsv:atColumn 1;
  oboe:ofCharacteristic [some_characteristic];
  oboe:usesStandard [some_standard] .

<ts0>
  a time:Instant;
  ccsv:atColumn 0 .
\end{lstlisting}
\noindent

\subsection{Domain ontology}

An Observation from the scientific point of view is defined as the act of observing some entity’s property and obtaining a measured value. Measurements are expressed using a specific standard (or unit). To have measurements properly annotated, we need a hierarchy of entities of interest with relevant properties, a hierarchy of entity characteristics with relevant properties, and a hierarchy of standards to be used in in measurements. The OBOE ontology already provides a hierarchy of a number of entities, characteristics and measurement units,  although it is focused on ecological studies. A domain ontology enables deep linking of the collected data to its metadata. For cities, Section 4 discusses our initial approach to this ontology in the context of a public transportation system.

\section{Use case: Fortaleza urban transportation system}

Our primary use cases focus on the city of Fortaleza, Brazil. The city of Fortaleza is the capital of the Ceará state in the Northeast region of Brazil with approximately 2.5 million inhabitants. Recently, the City Hall launched its smart city program called “Fortaleza Inteligente”. This project, coordinated by the Foundation for Science, Technology and Innovation of Fortaleza (CITINOVA), has as one of its main components the city open data portal, which contains a plethora of different datasets from a variety of public segments. We are collaborating with CITINOVA to use our approach in the city open data portal, in particular to represent data about the segment of urban mobility.

The first initiative was to model public transportation data. Fortaleza has deployed, on each of its running buses, GPS instruments capable of transmitting the current geographical position of the bus. GPS data is transmitted to a central server where they are stored in datasets that follow an ad hoc logic. The files should refer to a GPS position of a bus, for a given bus line and during a certain day. In other words GPS data is represented by a triplet bus-id, bus-line, date. However, this structure is not always followed. There are situations where the file becomes too large, and is then divided into two datasets. In addition, when a bus happens to be used in more than one line, then the same GPS receiver is used to collect more than one bus line’s position. This can all be represented in the same dataset, which again breaks the informal rule (bus-id, bus-line, date). Other variations exist, but the decisive factor is that none of these informal rules used for the construction of datasets were specified or clearly documented. They were discovered in several rounds of interviews with providers of data only when other users decided to explore the data. This makes the process not easily reusable or scalable.

\subsection{Building the Semantic Sensor Network}

In order to develop a proof of concept, we gathered three datasets from the Fortaleza Dados Abertos\footnote{http://dados.fortaleza.gov.br} portal that related to the public transportation system:

\begin{itemize}
\item Bus checkpoints: A list with all the bus checkpoints around the city. Each checkpoint is composed of its code, a name, a lat/long position and a radius around that lat/long to indicate the area the checkpoint that it covers.
\item Bus companies: A list with all the bus companies on contract with City Hall to provide that public service. Each company has a company id, a name and other information.
\item Bus fleet: A list with all the buses running in the city. Each bus has an id, a model, a maker, which company it belongs to and other information including plates, serial numbers and so forth.
\item GPS bus information for 2015/02: A report including when a specific bus entered and left a bus checkpoint.
\end{itemize}

We note that first three datasets can be viewed as metadata that plays a support role for the fourth dataset. This last dataset is the actual observed data that was collected during a data collection activity. We also note that information such as bus company and bus checkpoint provide contextual metadata. We have mapped the checkpoints and fleet datasets using the HASNetO-SC ontology:

\begin{itemize}
\item Checkpoints: By analysing the GPS dataset, we can infer that all the measurements contained inside it are coming from the checkpoints themselves. To reflect this, we then mapped all checkpoints to individual instances of the vstoi:Instrument as they are able to measure when a particular bus is entering or leaving its coverage area.
\item Checkpoints location: We have also inferred that the actual instruments identified in the previous item were deployed at the place described in it. For the purpose of this proof of concept, we defined that the place of deployment is an individual instance of hasneto-sc:RoadSegment, as all checkpoint’s lat/long locations are on roads.
\end{itemize}

The following is part of the Turtle describing our sensor network using HASNetO-SC.

\medskip

\noindent
{\it Serialized RDF Model of the Sensor Network}
\begin{lstlisting}[frame=single]
<checkpoint-1>
    a vstoi:Instrument ;
    rdfs:label "Dallas/Sobradinho/T F Paiva" .

<checkpoint-2>
    a vstoi:Instrument ;
    rdfs:label "Pedro Pereira/Imperador/T Goncalves" .    
...
<checkpoint-platform-1>
    a hasneto-sc:RoadSegment ;
    rdfs:label "Dallas/Sobradinho/T F Paiva" .
    geo:lat -3.79486600 ;
    geo:long -38.61625700 .

<checkpoint-platform-2>
    a hasneto-sc:RoadSegment ;
    rdfs:label "Pedro Pereira/Imperador/T Goncalves" .
    geo:lat -3.72791200 ;
    geo:long -38.53405200 .
...
\end{lstlisting}
\noindent

\subsection{Annotating the CSV dataset}

\begin{figure}
\centering
\includegraphics[height=2cm]{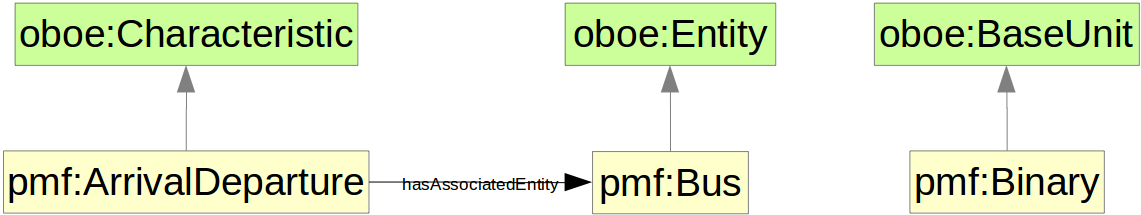}
\caption{Domain ontology for the bus use case.}
\label{domain-ontology}
\end{figure}

Once the semantic sensor network was ready, the next step was to annotate the CSV dataset containing the GPS bus information. The former dataset was a combined collection of measurements from different checkpoint instruments. To fully integrate this dataset with our process, we split it so that each resulting dataset contains information collected by a single instrument. To annotate each dataset, as mentioned before, we need a domain ontology that includes a hierarchy of entities, characteristics and units. For the purpose of the use case, we have developed a small ontology capable of describing the identified concepts, i.e., to state that we are measuring the occurrence (unit) of an arrival or departure (characteristic) of a bus (entity). As shown in Fig.~\ref{domain-ontology}, the ontology defines two classes under the namespace pmf (acronym for Prefeitura Municipal de Fortaleza - Fortaleza City Hall): pmf:BusArrivalDeparture that specializes oboe:Characteristic and pmf:Binary that specializes oboe:BaseUnit. Also, note that pmf:ArrivalDeparture characteristic has an associated entity pmf:Bus.

Given this, we annotated each dataset to build the following CCSV preamble.

\medskip

\noindent
{\it GPS bus dataset CCSV preamble}
\begin{lstlisting}[frame=single]
<pmf-kb>
  a ccsv:KnowledgeBase;
  ccsv:hasConnectionURL "http..."^^xsd:anyURI .

<deployment-checkpoint-1>
  a vstoi:Deployment;
  prov:startedAtTime "2015-02-01T00:00:00Z"^^xsd:dateTime;
  hasneto:hasDataCollection <datacollection-checkpoint-1> .

<datacollection-checkpoint-1>
  a hasneto:DataCollection; a time:Interval;
  prov:startedAtTime "2015-02-01T00:00:00Z"^^xsd:dateTime .

<gps-bus-information-checkpoint-1>
  a vstoi:Dataset;
  prov:wasGeneratedBy <datacollection-checkpoint-1> ;
  hasneto:hasMeasurementType <mt0> .

<mt0>
  a oboe:Measurement; a time:Instant;
  time:inDateTime <ts0>;
  ccsv:atColumn 1;
  oboe:ofCharacteristic pmf:ArrivalDeparture ;
  oboe:usesStandard pmf:Binary .

<ts0>
  a time:Instant; ccsv:atColumn 0 .
\end{lstlisting}
\noindent

The dataset was annotated to contain deployment information for that single instrument and also to specify that a new data collection activity was being initiated. It is also shown that this dataset has a measurement of characteristic pmf:ArrivalDepartureEvent in the pmf:Binary unit located on column number 1 that was taken on the timestamp specified at column 0.

\subsection{CCSV parsing and indexing}

In order to complete the connection between data and metadata described previously including the full integration with the semantic sensor network, we have deployed an instance of Apache SOLR\footnote{http://lucene.apache.org/solr/} to index and store both data and metadata. SOLR is a NoSQL database capable of indexing documents in a number of different formats including CSV, XML and JSON. SOLR document collections should include field definitions, including data types, for content that is to be indexed and stored. Inside SOLR, we have created two collections: (i) a metadata collection for storing the semantic sensor network, the domain ontology and the deployment metadata, and (ii) a measurement collection for storing the data itself.

The metadata collection is basically a triple store where the sensor network Turtle was loaded  along with the domain ontology described. The measurement collection is a regular SOLR collection with the following fields being indexed and stored:

\begin{itemize}
\item Entity, characteristic and unit
\item Location and timestamp
\item Measured value
\item Instrument
\item Data collection to which the measurement belongs
\end{itemize}

To automate the process of storing this information, we have developed an application called the CCSV-Loader\footnote{http://tw.rpi.edu/web/project/JeffersonProjectAtLakeGeorge/download/ccsv}. This loader performs, in order, the following tasks:

\begin{enumerate}
\item Extract the Turtle preamble of the CCSV dataset
\item Get instrument information based on deployment
\item Get platform information based on deployment
\item Generate a normalized CSV file with the extra columns containing metadata information
\item Index the normalized CSV file in the measurement collection
\end{enumerate}

\begin{figure}
\centering
\includegraphics[height=11.5cm]{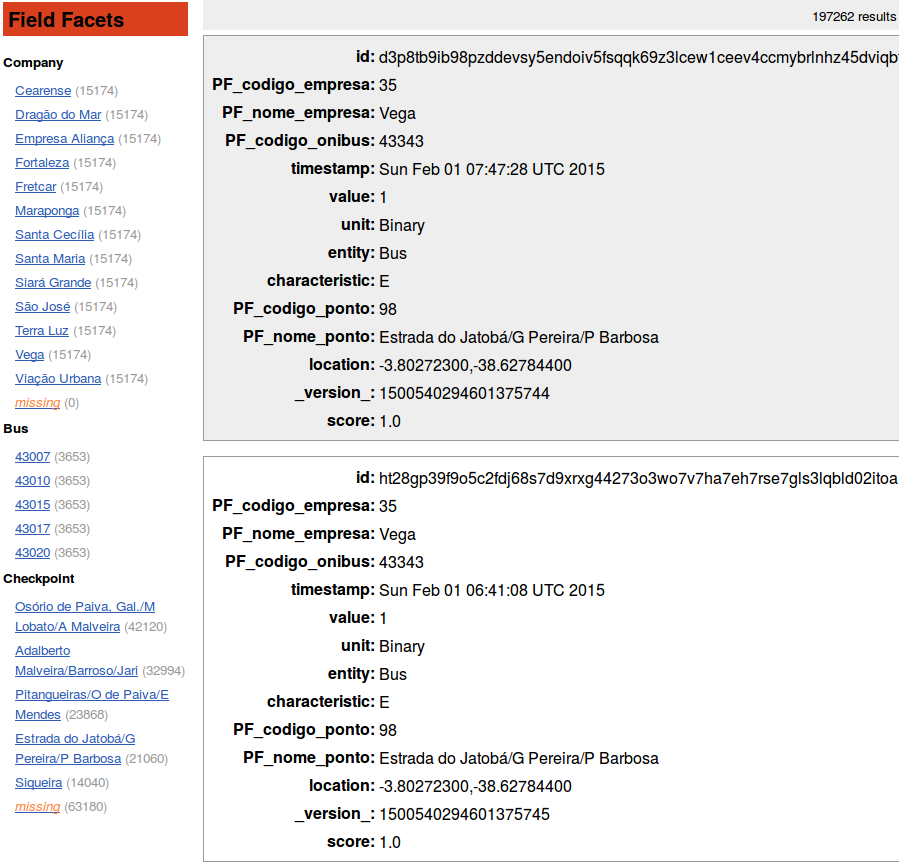}
\caption{Faceted-browsing of Fortaleza city transportation data.}
\label{fig-use-case-01}
\end{figure}

Once the data is indexed, we make use of SOLR faceted-search capabilities to present a navigation UI to the user. Fig.~\ref{fig-use-case-01} shows an example of that UI. On the right side, measurements are shown, based on the user current search parameters. On the left side, we have defined some fields to be faceted by SOLR to provide easy access to commonly requested content. For this use case, we have defined checkpoints as a field to be faceted so a user can select measurements coming from a specific checkpoint. Also, we added buses and companies to the facet field list, giving the ability for users to refine the search results by a bus or a company.

\section{Discussion}

This paper presented a streamlined process to collect, store and disseminate monitored data in an urban environment. The work was motivated by the challenges related to understanding, using, and integrating data from dataset portals based on current monitoring and publishing approaches. One issue we introduced is the boundary of a dataset as a natural context boundary vs. a technologically convenient boundary.  Our Fortaleza public transportation system use case showed that our approach enables users to have access not only to the sensor network metadata but also to metadata driven navigation of the data.  This combination we believe helps circumvent many of the challenges with understanding, using, and integrating data, even in settings where datasets have varied and potentially arbitrary boundaries and also in settings where metadata is limited.

In future work, we will articulate and implement additional use cases.  We are also working on tighter integration with the real time sensor network content in order to augment it using our metadata scheme thus further evaluating our HASNetO-SC ontology.  Additionally, we plan to  apply our approach in other Smart City initiatives. Our vision is to deploy our approach alongside open data portals, providing end users with a better understanding of complex monitored urban environment data.

\subsubsection{Acknowledgements.} The first author is supported by CNPq - Brazil.

\bibliography{biblio}

\begin{thebibliography}{10}
\providecommand{\url}[1]{\texttt{#1}}
\providecommand{\urlprefix}{URL }

\bibitem{compton_ssn_2012}
Compton, M., Barnaghi, P., Bermudez, L., García-Castro, R., Corcho, O., Cox,
  S., Graybeal, J., Hauswirth, M., Henson, C., Herzog, A., Huang, V., Janowicz,
  K., Kelsey, W.D., Le~Phuoc, D., Lefort, L., Leggieri, M., Neuhaus, H.,
  Nikolov, A., Page, K., Passant, A., Sheth, A., Taylor, K.: The {SSN} ontology
  of the {W}3c semantic sensor network incubator group. Web Semantics: Science,
  Services and Agents on the World Wide Web  17,  25--32 (Dec 2012)

\bibitem{compton_survey_2009}
Compton, M., Henson, C., Lefort, L., Neuhaus, H., Sheth, A.: A {Survey} of the
  {Semantic} {Specification} of {Sensors}. CEUR Workshop Proceedings pp. 17--32
  (Oct 2009)

\bibitem{cox_observations_2011}
Cox, S.: Observations and {Measurements} - {XML} {Implementation} (Mar 2011)

\bibitem{fensel_sesame-s:_2012}
Fensel, A., Tomic, S., Kumar, V., Stefanovic, M., Aleshin, S.V., Novikov, D.O.:
  {SESAME}-{S}: {Semantic} {Smart} {Home} {System} for {Energy} {Efficiency}.
  Informatik-Spektrum  36(1),  46--57 (Dec 2012)

\bibitem{fox_ontology-supported_2009}
Fox, P., McGuinness, D.L., Cinquini, L., West, P., Garcia, J., Benedict, J.L.,
  Middleton, D.: Ontology-supported scientific data frameworks: {The} {Virtual}
  {Solar}-{Terrestrial} {Observatory} experience. Computers \& Geosciences
  35(4),  724--738 (Apr 2009)

\bibitem{gao_semantic_2014}
Gao, F., Ali, M.I., Mileo, A.: Semantic {Discovery} and {Integration} of
  {Urban} {Data} {Streams}. In: Proceedings of the {Fifth} {Workshop} on
  {Semantics} for {Smarter} {Cities}. pp. 15--30. Riva del Garda, Italy (Oct
  2014)

\bibitem{giffinger_smart_2007}
Giffinger, R., Fertner, C., Kramar, H., Kalasek, R., Pichler-Milanovic, N.,
  Meijers, E.: Smart cities-{Ranking} of {European} medium-sized cities. Tech.
  rep., Vienna University of Technology (2007)

\bibitem{goodchild_citizens_2007}
Goodchild, M.F.: Citizens as sensors: the world of volunteered geography.
  GeoJournal  69(4),  211--221 (Nov 2007)

\bibitem{hendler_us_2012}
Hendler, J., Holm, J., Musialek, C., Thomas, G.: {US} {Government} {Linked}
  {Open} {Data}: {Semantic}.data.gov. IEEE Intelligent Systems  27(3),  25--31
  (May 2012)

\bibitem{kuhn_functional_2009}
Kuhn, W.: A {Functional} {Ontology} of {Observation} and {Measurement}. In:
  Janowicz, K., Raubal, M., Levashkin, S. (eds.) {GeoSpatial} {Semantics}, pp.
  26--43. No. 5892 in Lecture {Notes} in {Computer} {Science}, Springer Berlin
  Heidelberg (2009)

\bibitem{lopez_queriocity:_2012}
Lopez, V., Kotoulas, S., Sbodio, M.L., Stephenson, M., Gkoulalas-Divanis, A.,
  Aonghusa, P.M.: {QuerioCity}: {A} {Linked} {Data} {Platform} for {Urban}
  {Information} {Management}. In: Cudré-Mauroux, P., Heflin, J., Sirin, E.,
  Tudorache, T., Euzenat, J., Hauswirth, M., Parreira, J.X., Hendler, J.,
  Schreiber, G., Bernstein, A., Blomqvist, E. (eds.) The {Semantic} {Web} –
  {ISWC} 2012, pp. 148--163. No. 7650 in Lecture {Notes} in {Computer}
  {Science}, Springer Berlin Heidelberg (Jan 2012)

\bibitem{madin_ontology_2007}
Madin, J., Bowers, S., Schildhauer, M., Krivov, S., Pennington, D., Villa, F.:
  An ontology for describing and synthesizing ecological observation data.
  Ecological Informatics  2(3),  279--296 (Oct 2007)

\bibitem{pinheiro_hasneto_2015}
Pinheiro, P., McGuinness, D.L., Santos, H.: Human-{Aware} {Sensor} {Network}
  {Ontology}: Semantic {Support} for {Empirical} {Data} {Collection}. In:
  Proceedings of the 5th Workshop on Linked Science. Bethlehem, PA, USA (2015)

\bibitem{probst_ontological_2006}
Probst, F.: Ontological {Analysis} of {Observations} and {Measurements}. In:
  Raubal, M., Miller, H.J., Frank, A.U., Goodchild, M.F. (eds.) Geographic
  {Information} {Science}, pp. 304--320. No. 4197 in Lecture {Notes} in
  {Computer} {Science}, Springer Berlin Heidelberg (2006)

\bibitem{quine_stimulus_1995}
Quine, W.V.O.: From {Stimulus} to {Science}. Harvard University Press (1995)

\bibitem{rew_netcdf:_1990}
Rew, R., Davis, G.: {NetCDF}: an interface for scientific data access. IEEE
  Computer Graphics and Applications  10(4),  76--82 (Jul 1990)

\bibitem{shadbolt_linked_2012}
Shadbolt, N., O'Hara, K., Berners-Lee, T., Gibbins, N., Glaser, H., Hall, W.,
  schraefel, m.: Linked {Open} {Government} {Data}: {Lessons} from
  {Data}.gov.uk. IEEE Intelligent Systems  27(3),  16--24 (May 2012)

\bibitem{stasch_stimulus-centric_2009}
Stasch, C., Janowicz, K., Bröring, A., Reis, I., Kuhn, W.: A
  {Stimulus}-{Centric} {Algebraic} {Approach} to {Sensors} and {Observations}.
  In: Trigoni, N., Markham, A., Nawaz, S. (eds.) {GeoSensor} {Networks}, pp.
  169--179. No. 5659 in Lecture {Notes} in {Computer} {Science}, Springer
  Berlin Heidelberg (2009)

\bibitem{stavropoulos_bonsai:_2012}
Stavropoulos, T.G., Vrakas, D., Vlachava, D., Bassiliades, N.: {BOnSAI}: {A}
  {Smart} {Building} {Ontology} for {Ambient} {Intelligence}. In: Proceedings
  of the 2Nd {International} {Conference} on {Web} {Intelligence}, {Mining} and
  {Semantics}. pp. 30:1--30:12. {WIMS} '12, ACM, New York, NY, USA (2012)

\bibitem{usbeck_combining_2014}
Usbeck, R.: Combining {Linked} {Data} and {Statistical} {Information}
  {Retrieval}. In: Presutti, V., d’Amato, C., Gandon, F., d’Aquin, M.,
  Staab, S., Tordai, A. (eds.) The {Semantic} {Web}: {Trends} and {Challenges},
  pp. 845--854. No. 8465 in Lecture {Notes} in {Computer} {Science}, Springer
  International Publishing (Jan 2014)

\end{thebibliography}

\end{document}